\PassOptionsToPackage{unicode}{hyperref}
\PassOptionsToPackage{hyphens}{url}
\documentclass[
]{article}
\usepackage{xcolor}
\usepackage{amsmath,amssymb}
\usepackage{fancyhdr}
\usepackage{iftex}
\ifPDFTeX
  \usepackage[T1]{fontenc}
  \usepackage[utf8]{inputenc}
  \usepackage{textcomp} 
\else 
  \usepackage{unicode-math} 
  \defaultfontfeatures{Scale=MatchLowercase}
  \defaultfontfeatures[\rmfamily]{Ligatures=TeX,Scale=1}
\fi
\usepackage{lmodern}
\ifPDFTeX\else
\fi
\IfFileExists{upquote.sty}{\usepackage{upquote}}{}
\IfFileExists{microtype.sty}{
  \usepackage[]{microtype}
  \UseMicrotypeSet[protrusion]{basicmath} 
}{}
\makeatletter
\@ifundefined{KOMAClassName}{
  \IfFileExists{parskip.sty}{%
    \usepackage{parskip}
  }{
    \setlength{\parindent}{0pt}
    \setlength{\parskip}{6pt plus 2pt minus 1pt}}
}{
  \KOMAoptions{parskip=half}}
\makeatother
\usepackage{longtable,booktabs,array}
\usepackage{multirow}
\usepackage{calc} 
\usepackage{etoolbox}
\makeatletter
\patchcmd\longtable{\par}{\if@noskipsec\mbox{}\fi\par}{}{}
\makeatother
\IfFileExists{footnotehyper.sty}{\usepackage{footnotehyper}}{\usepackage{footnote}}
\makesavenoteenv{longtable}
\usepackage{graphicx}
\makeatletter
\newsavebox\pandoc@box
\newcommand*\pandocbounded[1]{
  \sbox\pandoc@box{#1}%
  \Gscale@div\@tempa{\textheight}{\dimexpr\ht\pandoc@box+\dp\pandoc@box\relax}%
  \Gscale@div\@tempb{\linewidth}{\wd\pandoc@box}%
  \ifdim\@tempb\p@<\@tempa\p@\let\@tempa\@tempb\fi
  \ifdim\@tempa\p@<\p@\scalebox{\@tempa}{\usebox\pandoc@box}%
  \else\usebox{\pandoc@box}%
  \fi%
}
\def\fps@figure{htbp}
\makeatother
\usepackage{bookmark}
\IfFileExists{xurl.sty}{\usepackage{xurl}}{} 
\hypersetup{
  hidelinks,
  pdfcreator={LaTeX via pandoc}}

\usepackage{authblk}
\author[1,2]{Zhoupeng Shou}
\author[1]{Zhiqiang You}
\author[1]{Fang Wang}
\author[3*]{Haibo Liu}
\affil[1]{NoDesk AI, Hangzhou, China}
\affil[2]{Zhejiang University, Hangzhou, China}
\affil[3]{Independent Researcher, Hangzhou, China}
\affil[*]{Corresponding author: Haibo Liu (haiboliu2025@gmail.com)}
\title{CogGuide: Human-Like Guidance for Zero-Shot Omni-Modal Reasoning}
\date{}

\begin{document}
\maketitle

\emph{\textbf{Abstract---}}Addressing shortcut reliance and limited contextual understanding in cross-modal reasoning, we introduce a zero-shot omni-modal reasoning component inspired by human-like cognition, realized as a plug-and-play intent-sketch pipeline with three serial modules: Intent Perceiver, Policy Generator and Strategy Selector, that models an understand-plan-select cognitive process. By producing and filtering lightweight intent sketch strategies to guide reasoning, the method requires no parameter fine-tuning and enables cross-model transfer through in-context engineering. An information-theoretic analysis demonstrates that this process reduces conditional entropy and improves information utilization efficiency. Extensive experiments on IntentBench, WorldSense, and Daily-Omni confirm the method’s generality and robustness: the full three-module design consistently outperforms strong baselines across diverse reasoning engines and pipeline settings, with maximum gains of +9.51 pp and a relative improvement of 20.04\%. These results demonstrate the practical value and portability of the proposed intent sketch reasoning component for zero-shot omni-modal reasoning.

\emph{\textbf{Index Terms---}}LLM, Omni-Modal, Intent Sketch,
Information Entropy

\includegraphics[width=4.9in,height=1.45486in]{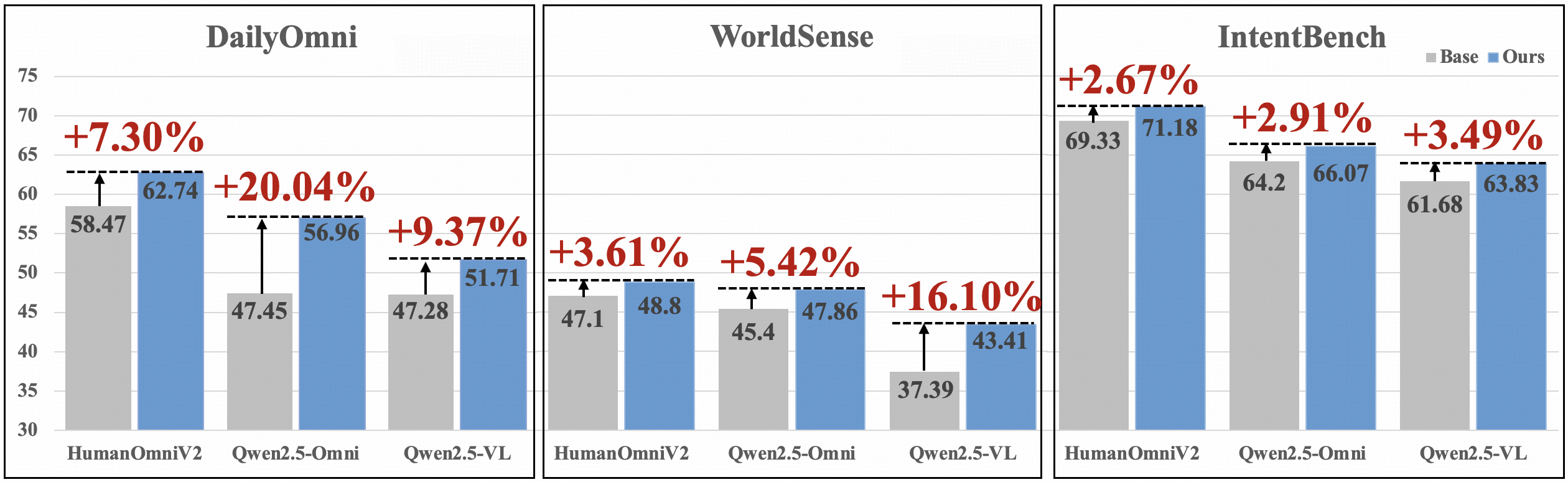}

\textbf{Figure 1.} Relative Percentage Improvements over Baselines by
the Proposed Method on IntentBench, WorldSense, and Daily-Omni.

\section{\texorpdfstring{
INTRODUCTION}{ INTRODUCTION}}\label{introduction}

In recent years, multimodal large language models (MLLMs) have been
propelling artificial intelligence from purely textual understanding
toward comprehensive understanding and reasoning across ``all
modalities,'' including vision, speech, and video. In 2023, OpenAI
released GPT-4, which for the first time introduced visual input
capabilities, enabling LLMs to parse images and perform reasoning
{[}1{]}. Since then, numerous researchers have begun developing and
applying multimodal large models {[}2{]}, {[}3{]}. Against this
backdrop, tasks such as video question answering impose higher demands
on dynamic-scene comprehension and temporal reasoning{[}4{]}, motivating
research to unify, align, and fuse vision, language, and audio in order
to enhance cross-modal understanding and reasoning {[}5{]}.
Consequently, multimodal large models have expanded from images to video
and audio, evolving toward a general ``audiovisual omnipotent''
architecture that strengthens world representations while broadening
human--computer interaction {[}6{]}.

However, existing multimodal models still have shortcomings in complex
reasoning scenarios. Even with massive parameters and multimodal
pre-training, they continue to face challenges such as insufficient
global contextual understanding and ``shortcut'' reasoning in complex
tasks {[}7{]}. Models often over-rely on local or single-modality cues
while overlooking critical cross-modal information, leading to outputs
that deviate from human intent {[}8{]}. On the other hand, even when
employing prompting strategies such as Chain-of-Thought, current
multimodal large models remain notably limited on multi-step cross-modal
reasoning tasks {[}9{]}. Constructing reasoning chains via reinforcement
learning may cause models to acquire ``shortcut'' strategies, thereby
reducing generalization {[}10{]}. These phenomena indicate that relying
solely on the model's inherent reasoning ability and simple prompts is
still insufficient to align the model with human intent.

To address these issues, researchers have begun incorporating ``intent''
as a mediating variable connecting user queries and cross-modal evidence
into the multimodal reasoning loop: one line of work introduces explicit
intent labels or scene purposes in multimodal question answering and
video understanding to constrain candidate answers and the scope of
reasoning {[}7{]}; another employs instruction tuning and templated
prompts to explicitly declare desired behaviors on the input side,
thereby implicitly aligning reasoning goals {[}11{]}; still others
construct ``intent-conditioned'' retrieval--reasoning pipelines or
agents so that evidence selection and reasoning steps are driven by the
current intent {[}12{]} ; there are also methods that adopt text-guided
multimodal fusion to assist multimodal intent understanding {[}13{]}.
Meanwhile, attempts to exploit the temporal and event structures of
audio--video to infer latent intent and filter irrelevant cues are
increasing {[}14{]}, {[}15{]}. However, these methods have certain
limitations: they rely on dense annotation and task-specific training,
making zero-shot transfer difficult {[}7{]}, {[}16{]}, or they treat
intent as static labels or prompt fragments without externalizing it
into generable, assessable, and selectable strategies, thereby failing
to stably suppress ``shortcut'' reasoning and local biases {[}8{]},
{[}11{]}.

Recently, studies have begun to explore multimodal reasoning that
simulates human ``sketches of thought.'' For example, the
Sketch-of-Thought framework {[}17{]} introduces a human-like cognitive
reasoning paradigm that maintains reasoning accuracy while reducing
verbose intermediate reasoning; the Machine Mental Imagery approach
{[}18{]} enables models to incorporate latent visual ``imagination''
representations during reasoning, simulating the human process of using
mental imagery to aid reasoning, thereby improving reasoning efficiency
and understanding of complex scenes. Inspired by such research, we
propose a human-like cognitive intent-sketch reasoning component that
modularly enhances multimodal intent recognition and reasoning
performance. The core idea is to decompose and recombine the context of
complex problems and to embed human cognitive processes into model
reasoning through explicit strategy-planning prompts (i.e., the ``intent
sketch'' mechanism), thereby strengthening the model's understanding of
multimodal information and latent intent. Specifically, we propose a zero-shot omni-modal reasoning component inspired by human-like cognition. The component decomposes context through a plug-and-play intent-sketch pipeline: Intent Perceiver, Policy Generator and Strategy Selector, which infer textual intent from video and audio, generate candidate policies, and select an optimal strategy; a reasoning LLM then conditions on the selected strategy to produce the final answer via prompt composition. 
On multiple challenging multimodal reasoning benchmarks (e.g., IntentBench
{[}7{]}, WorldSense{[}14{]} and Daily-Omni{[}15{]}), using HumanOmniV2, Qwen2.5-Omni, Qwen2.5-VL, respectively, serving as a unified comparison. Our method surpasses the base model on all benchmarks (highest relative increase: 20.04\%). Meanwhile, in terms of “effectiveness vs. cost” compared with training-based methods, we maintain zero training overhead and low latency, resulting in better overall cost and deployment timelines. 

The main contributions: (1) a zero-shot omni-modal reasoning component that contains a plug-and-play intent-sketch pipeline, deliver immediate accuracy gains and cross-model generality; (2) an information-entropy analysis that explains how strategy prompts reduce decision uncertainty; (3) present comprehensive omni-modal experiments, including ablations and reasoning-engine swaps, that validate the effectiveness and practicality of in-context strategy prompting on the three benchmarks above.

\includegraphics[width=5in,height=2.45833in]{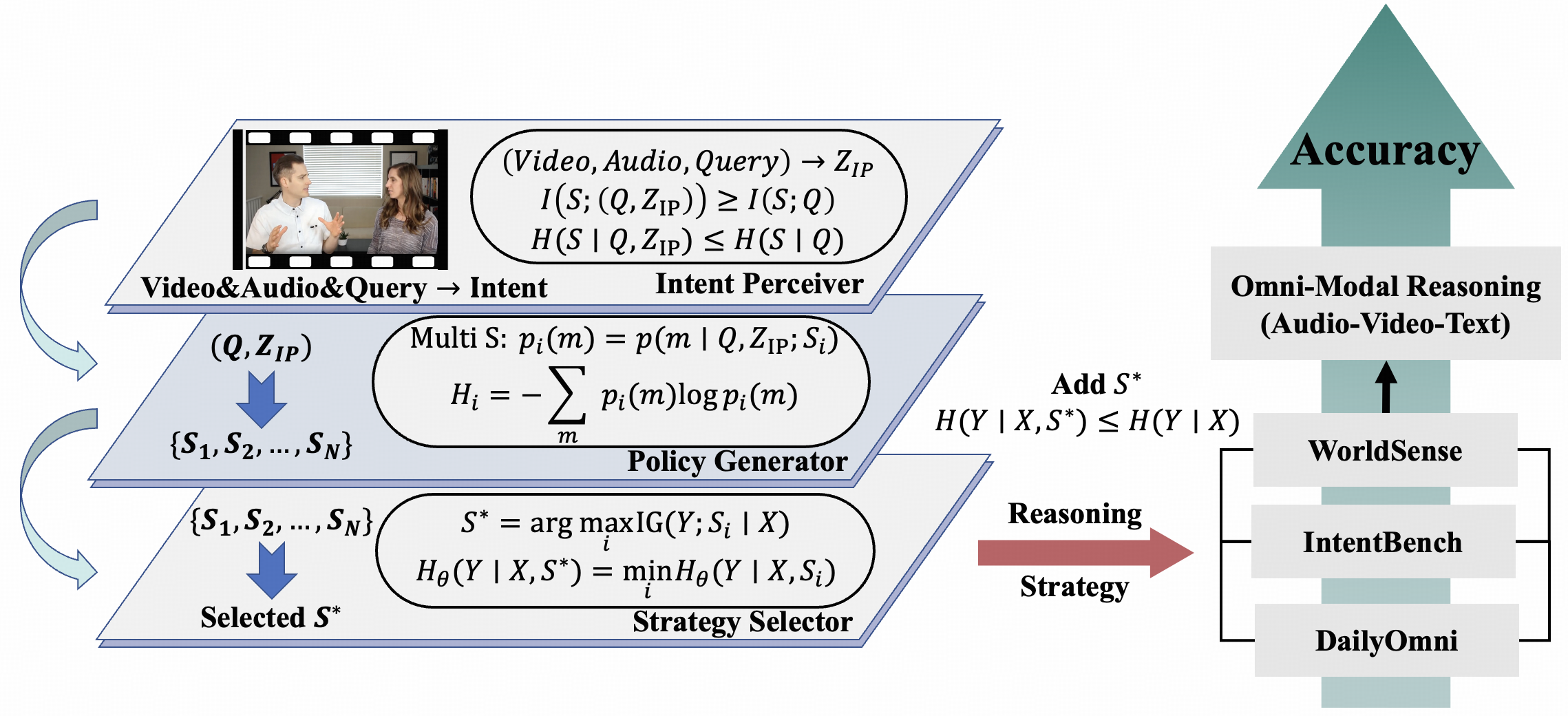}

\textbf{Figure 2.} Zero-shot omni-modal reasoning component: plug-and-play intent–sketch pipeline and information-entropy analysis

\section{\texorpdfstring{METHOD}{METHOD}}\label{method}

The plug-and-play intent–sketch pipeline contains three serial modules: \emph{Intent Perceiver}, \emph{Policy Generator}, and \emph{Strategy Selector}, operating on omni-modal input \(X=(V,A,Q)\) where \(V\) is video, \(A\) is audio, and \(Q\) is the textual query. The Intent Perceiver maps \(X\) to \(Z_{\mathrm{IP}}\) to reduce planning uncertainty; the Policy Generator yields \(N\) semantically diverse policies (strategy sketches); the Strategy Selector applies extra conditioning \(C\) and selects \(S^{*}\) to drive the final reasoning. This design mirrors the human routine of \emph{understand–plan–select}, yielding a clearer reasoning trajectory and improved answer generation.

\subsection{2.1\quad Intent Perceiver: Intent Representation for Strategy Generation}\label{sec:ip}

Given \(X=(V,A,Q)\), the module outputs an intent representation \(Z_{\mathrm{IP}} = f(X)\) that fuses cross-modal cues with the query. From an information-theoretic view, using \(Z_{\mathrm{IP}}\) to condition strategy generation provides nonnegative information gain:
\begin{equation}
\mathrm{MI}\!\left(S;\,(Q,Z_{\mathrm{IP}})\right) - \mathrm{MI}(S;Q)
= \mathrm{MI}\!\left(S;Z_{\mathrm{IP}} \mid Q\right) \ge 0,
\label{eq:mi_gain}
\end{equation}
which is equivalent to \(H(S\mid Q,Z_{\mathrm{IP}})\le H(S\mid Q)\). By focusing on answer-relevant omni-modal evidence, \(Z_{\mathrm{IP}}\) reduces ambiguity at problem understanding and effectively lowers the initial conditional uncertainty \(H(Y\mid X)\).

\subsection{2.2\quad Policy Generator: Omni-Policy Generation Based on Semantic Entropy}\label{sec:pg}

Conditioned on \((Q,Z_{\mathrm{IP}})\), a frozen LLM (prompted as a policy provider) generates \(N\) candidate sketches \(\{S_1,\ldots,S_N\}\) that specify lines of reasoning without producing the answer. Let \(\mathcal{M}\) denote semantic equivalence classes over policies and \(p(m\mid Q,Z_{\mathrm{IP}})\) the posterior over classes. We define the semantic entropy
\begin{equation}
H_{\mathrm{sem}}(S\mid Q,Z_{\mathrm{IP}})
= -\sum_{m\in\mathcal{M}} p(m\mid Q,Z_{\mathrm{IP}})\,\log p(m\mid Q,Z_{\mathrm{IP}}),
\label{eq:sem_entropy}
\end{equation}
and the gain brought by intent conditioning:
\begin{equation}
\mathrm{MI}(S;Z_{\mathrm{IP}}\mid Q)
= H_{\mathrm{sem}}(S\mid Q)-H_{\mathrm{sem}}(S\mid Q,Z_{\mathrm{IP}})\ \ge\ 0.
\label{eq:sem_gain}
\end{equation}

To balance single-sketch clarity and set-level coverage, for each candidate \(S_i\) let \(p_i(m)=p(m\mid Q,Z_{\mathrm{IP}};S_i)\) and define
\begin{equation}
H_i = -\sum_{m} p_i(m)\,\log p_i(m).
\label{eq:single_entropy}
\end{equation}
For the mixture \(\bar p(m)=\tfrac{1}{N}\sum_{i=1}^{N} p_i(m)\) with entropy \(H(\bar p)=-\sum_m \bar p(m)\log \bar p(m)\), we optimize
\begin{equation}
\max_{S_1,\ldots,S_N}\ \ H(\bar p)\;-\; \alpha\,\frac{1}{N}\sum_{i=1}^{N} H_{\mathrm{sem}}(S_i\mid Q,Z_{\mathrm{IP}})
\;+\; \gamma\,\mathrm{Div}(S_1,\ldots,S_N),
\label{eq:set_objective}
\end{equation}
where \(\alpha,\gamma>0\) are weights, and \(\mathrm{Div}(\cdot)\) can be defined via pairwise distances in semantic similarity between candidates to encourage complementary policies with different emphases, such as ``evidence-first,'' ``temporal/causal-first,'' or ``cross-modal-alignment-first.''

\subsection{2.3\quad Strategy Selector: Selection via Minimum Conditional Entropy / Bayesian Risk}\label{sec:ss}

Given \(\{S_1,\ldots,S_N\}\), the model (prompted as a strategy evaluator) selects \(S^{*}\) based on the input \(X\) and task-specific conditioning \(C\) to align the choice with real-world constraints. Let \(p_{\theta}(Y\mid X,S_i,C)\) be the answer posterior under \(S_i\) with extra conditioning \(C\). In implementation, we obtain \(p_{\theta}(Y\mid X,S_i,C)\) by prompting the LLM with \((X,S_i,C)\) and computing answer-slot likelihoods, so the added \(C\) induces strategy-dependent posteriors. Under \(0\!-\!1\) loss,
\begin{equation}
S^{*} = \arg\min_{i} R_{0\text{--}1}(S_i) 
= \arg\max_{i}\ \max_{Y}\ p_{\theta}(Y\mid X,S_i,C).
\label{eq:bayes_risk}
\end{equation}
Selection maximizes information gain:
\begin{equation}
IG(Y;S_i\mid X) = H_{\theta}(Y\mid X) - H_{\theta}(Y\mid X,S_i,C),
\label{eq:ig_def}
\end{equation}
thus
\begin{equation}
S^{*}=\arg\max_{i}\ IG(Y;S_i\mid X) 
= \arg\min_{i}\ H_{\theta}(Y\mid X,S_i,C).
\label{eq:ig_argmin}
\end{equation}
Because candidates are complementary,
\begin{equation}
H_{\theta}(Y\mid X,S^{*}) 
= \min_{i}\ H_{\theta}(Y\mid X,S_i,C)
\ \le\ \frac{1}{N}\sum_{i=1}^{N} H_{\theta}(Y\mid X,S_i,C).
\label{eq:min_vs_mean}
\end{equation}
In practice, exploring a diverse and complementary set of strategies increases the chance that at least one \(S_i\) substantially reduces uncertainty, making lower overall uncertainty more likely. This implies a lower expected conditional entropy—and, via a Fano-type bound, a lower achievable error bound; in particular, reducing \(H_{\theta}(Y\mid X,S^{*},C)\) lowers an upper bound on the minimum error rate under \(0\!-\!1\) loss. Consequently, subsequent reasoning based on \(S^{*}\) integrates omni-modal evidence along a path of higher confidence, enhancing the reliability of answer generation.

\subsection{2.4\quad A Unified Information-Theoretic Framework of Uncertainty Reduction}\label{sec:unified}

This section formalizes our pipeline as an information-theoretic mechanism for uncertainty reduction. We begin with the basic theorem of conditioning:
\begin{equation}
H(X\mid Y)\ \le\ H(X).
\label{eq:conditioning}
\end{equation}
Treating the selected strategy \(S^{*}\) as an observable planning variable, and letting \(C\) denote the extra task conditioning used to choose \(S^{*}\), we model the answer posterior as \(p(Y\mid X,I,S,C)\), where \(I\equiv Z_{\mathrm{IP}}\). By the conditioning-reduces-entropy theorem, we obtain the following monotone contraction chain:
\begin{equation}
H(Y\mid X)\ \ge\ H(Y\mid X,I)\ \ge\ H(Y\mid X,I,S)\ \ge\ H(Y\mid X,I,S,C).
\label{eq:chain_contraction}
\end{equation}
This captures how intent \(I\), the multi-policy set \(S\), and the selected strategy \(S^{*}\) progressively reduce the conditional uncertainty of \(Y\).

From the perspective of the Data Processing Inequality (DPI), if \(X\!\to\! I\!\to\! S\!\to\! S^{*}\) forms a Markov chain, then
\begin{equation}
I(X;S^{*})\ \le\ I(X;S)\ \le\ I(X;I)\ \le\ I(X;X),
\label{eq:dpi}
\end{equation}
which implies that any representation derived from \(X\) must lose some information about \(X\) itself. However, in inference, we do not replace \(X\) with \(S^{*}\); instead, we use \(S^{*}\) as auxiliary information \emph{conditioned on} \(X\):
\begin{equation}
H(Y\mid X,S^{*})\ <\ H(Y\mid X),
\label{eq:strict_reduction}
\end{equation}
with strict inequality when \(I(Y;S^{*}\mid X)>0\). This reduction occurs because \(S^{*}\), although derived from \(X\), is constructed through a process that incorporates task-specific guidance (e.g., evidence-first, causal-first, cross-modal alignment). These human-informed design choices embed additional relevant cues into \(S^{*}\), enabling it to provide new information about \(Y\) that is not directly accessible from \(X\) alone. Thus, conditioning on \(S^{*}\) enhances prediction accuracy, leading to the entropy reduction in \eqref{eq:strict_reduction} and contributing to the overall uncertainty contraction in \eqref{eq:chain_contraction}.

\begin{longtable}[]{@{}
  >{\arraybackslash}m{(\linewidth - 0\tabcolsep) * \real{1.0}}@{}}
\toprule\noalign{}
\begin{minipage}[b]{\linewidth}\centering
\textbf{Algorithm 1: Pseudocode for intent sketch reasoning component}
\end{minipage} \\
\midrule\noalign{}
\endhead
\bottomrule\noalign{}
\endlastfoot

\(\text{\textbf{Input }} X=(V,A,Q),\ k;\ \text{\textbf{Output }} y, R\) \\
\(I \gets \mathrm{IntentPerceiver}(X)\) \\
\(S \gets \mathrm{StrategyGenerator}(I,Q,k)=\{s_{1},\ldots,s_{k}\}\) \\
\(\text{\textbf{function}}\ \mathrm{POSTERIOR\_AND\_ENTROPY}(X,s)\) \\
\(\quad P_{Y} \leftarrow \mathrm{ReasoningEngine}.\mathrm{posterior}(Y \mid X,s)\) \\
\(\quad H \leftarrow \mathrm{Entropy}(P_{Y})\) \\
\(\quad \text{\textbf{return}}\ (P_{Y}, H)\) \\
\(\mathrm{best\_H}\gets +\infty;\ \mathrm{best\_s}\gets \emptyset;\ \mathrm{post\_best}\gets \emptyset\) \\
\(\text{\textbf{for each}}~ s~ \text{\textbf{in}}~ S~ \text{\textbf{do}}\) \\
\(\quad P_{Y}, H \leftarrow \mathrm{POSTERIOR\_AND\_ENTROPY}(X,s)\) \\
\(\quad \text{if } H < \mathrm{best\_H}:\) \\
\(\qquad \mathrm{best\_H}\gets H;\ \mathrm{best\_s}\gets s;\ \mathrm{post\_best}\gets P_{Y}\) \\
\(\text{\textbf{end for}}\) \\
\(s^{\ast} \gets \mathrm{best\_s}\) \\
\((R, y) \leftarrow \mathrm{ReasoningEngine}.\mathrm{solve\_with\_strategy}(X, s^{\ast})\) \\
\(\text{\textbf{return }} y, R\) \\

\end{longtable}

\section{\texorpdfstring{ EXPERIMENTS AND
RESULTS}{ EXPERIMENTS AND RESULTS}}\label{experiments-and-results}

\subsection{3.1 Experimental Settings}\label{experimental-settings}

We evaluate the method on three omni-modal reasoning benchmarks: IntentBench (omni-intent understanding), WorldSense (audio-video collaborative analysis), and Daily-Omni (daily-life scenarios). Accuracy (\%) is the metric. We compare baseline models (no strategy prompts) with our approach that adds three plug-in modules—Intent Perceiver, Policy Generator, and Strategy Selector—whose output is passed to the reasoning engine (Table I).

All experiments follow a zero-shot setting (no fine-tuning; zero-shot prompts only). We use three reasoning engines: HumanOmniV2, Qwen2.5-Omni, and Qwen2.5-VL. The pipeline modules are instantiated with four pretrained LLMs: closed-source GPT-4o and Doubao-Seed-1.6, and open-source GLM-4.5 and Qwen3. To quantify module contributions, we conduct ablations (Table II). Each setting is evaluated over the Cartesian product of the four pipeline LMs and the three reasoning engines, with all other hyperparameters and inference configurations held constant for fair comparison. Through these combinations we measure the overall gain of the integrated system and the contribution of each module.

\textbf{Table 1.} Summary of Models, Roles, and Scales (``a/b'' denotes
total / activated parameters, for Mixture-of-Experts models)

\begin{longtable}[]{@{}
  >{\centering\arraybackslash}m{(\linewidth - 4\tabcolsep) * \real{0.28}}
  >{\centering\arraybackslash}m{(\linewidth - 4\tabcolsep) * \real{0.52}}
  >{\centering\arraybackslash}m{(\linewidth - 4\tabcolsep) * \real{0.2}}@{}}
\toprule\noalign{}
\begin{minipage}[b]{\linewidth}\centering
\textbf{Model}
\end{minipage} & \begin{minipage}[b]{\linewidth}\centering
\textbf{Role}
\end{minipage} & \begin{minipage}[b]{\linewidth}\centering
\textbf{Parameter Scale}
\end{minipage} \\
\midrule\noalign{}
\endhead
\bottomrule\noalign{}
\endlastfoot
\textbf{HumanOmniV2{[}7{]}} & Reasoning Engine & 7B \\
\textbf{Qwen2.5-Omni} & Reasoning Engine & 7B \\
\textbf{Qwen2.5-VL} & Reasoning Engine & 7B \\
\textbf{GPT-4o} & Policy Generator/Strategy Selector & Large
Closed-Source Model \\
\textbf{GLM-4.5} & Policy Generator/Strategy Selector & 355B/32B \\
\textbf{Doubao-Seed-1.6} & Policy Generator/Strategy Selector & Large
Closed-Source Model \\
\textbf{Qwen3} & Policy Generator/Strategy Selector & 235B/22B \\
\textbf{Qwen2.5-VL-32B} & Intent Perceiver & 32B \\
\textbf{GLM-4.5V} & Intent Perceiver & 106B/12B \\
\end{longtable}

\textbf{Table 2.} Experimental Configurations: Three-Module Settings

\begin{longtable}[]{@{}
  >{\centering\arraybackslash}m{0.25\linewidth}
  >{\centering\arraybackslash}m{0.29\linewidth}
  >{\centering\arraybackslash}m{0.18\linewidth}
  >{\centering\arraybackslash}m{0.18\linewidth}@{}}
\toprule
\textbf{Experiment ID} & \textbf{Intent Perceiver} & \textbf{Policy Generation} & \textbf{Strategy Selection} \\
\midrule
\endhead
\textbf{CG\_Qwen} & Qwen2.5-VL-32B & 3 & On \\
\textbf{CG\_GLM}  & GLM-4.5V        & 3 & On \\
\textbf{Abl\_NI}  & No              & 3 & On \\
\textbf{Abl\_SP}  & Qwen2.5-VL-32B  & 1 & On \\
\textbf{BaseLine} & No              & No & No \\
\bottomrule
\end{longtable}

\noindent\footnotesize
\textit{Notes. CG\_Qwen: Full pipeline, use Qwen as Intent model. CG\_GLM: Full pipeline, use GLM as Intent model. Abl\_NI: Remove the Intent module. Abl\_SP: Change policy generation to single policy. BaseLine: No front-end pipeline.}

\subsection{3.2 Main Results}\label{main-results}

\textbf{Table 3.} IntentBench dataset: accuracy (\%) across combinations
of reasoning models and Pipeline module models

\begin{longtable}[]{@{}
  >{\centering\arraybackslash}m{(\linewidth - 12\tabcolsep) * \real{0.2144}}
  >{\centering\arraybackslash}m{(\linewidth - 12\tabcolsep) * \real{0.0116}}
  >{\centering\arraybackslash}m{(\linewidth - 12\tabcolsep) * \real{0.2023}}
  >{\centering\arraybackslash}m{(\linewidth - 12\tabcolsep) * \real{0.1429}}
  >{\centering\arraybackslash}m{(\linewidth - 12\tabcolsep) * \real{0.1429}}
  >{\centering\arraybackslash}m{(\linewidth - 12\tabcolsep) * \real{0.1429}}
  >{\centering\arraybackslash}m{(\linewidth - 12\tabcolsep) * \real{0.1429}}@{}}
\toprule\noalign{}
\begin{minipage}[b]{\linewidth}\centering
Pipeline
\end{minipage} &
\multicolumn{2}{>{\centering\arraybackslash}m{(\linewidth - 12\tabcolsep) * \real{0.2139} + 2\tabcolsep}}{%
\begin{minipage}[b]{\linewidth}\centering
Reasoning Model

(Baseline)
\end{minipage}} & \begin{minipage}[b]{\linewidth}\centering
CG\_Qwen

\end{minipage} & \begin{minipage}[b]{\linewidth}\centering
CG\_GLM

\end{minipage} & \begin{minipage}[b]{\linewidth}\centering
Abl\_NI
\end{minipage} & \begin{minipage}[b]{\linewidth}\centering
Abl\_SP

\end{minipage} \\
\midrule\noalign{}
\endhead
\bottomrule\noalign{}
\endlastfoot
\multicolumn{2}{@{}>{\centering\arraybackslash}m{(\linewidth - 12\tabcolsep) * \real{0.2260} + 2\tabcolsep}}{%
GPT-4o} &
\multirow{4}{=}{\centering\arraybackslash \begin{minipage}[t]{\linewidth}\centering
HumanOmniV2\\
(69.33) {[}7{]}\strut
\end{minipage}} & 70.86 & 70.47 & 70.45 & 70.09 \\
\multicolumn{2}{@{}>{\centering\arraybackslash}m{(\linewidth - 12\tabcolsep) * \real{0.2260} + 2\tabcolsep}}{%
GLM-4.5} & & 71.07 & 70.87 & 70.51 & 70.27 \\
\multicolumn{2}{@{}>{\centering\arraybackslash}m{(\linewidth - 12\tabcolsep) * \real{0.2260} + 2\tabcolsep}}{%
Doubao-Seed-1.6} & & 70.92 & 70.72 & 70.06 & 69.74 \\
\multicolumn{2}{@{}>{\centering\arraybackslash}m{(\linewidth - 12\tabcolsep) * \real{0.2260} + 2\tabcolsep}}{%
Qwen3} & & 71.18 & 70.9 & 70.82 & 69.96 \\
\multicolumn{2}{@{}>{\centering\arraybackslash}m{(\linewidth - 12\tabcolsep) * \real{0.2260} + 2\tabcolsep}}{%
GPT-4o} &
\multirow{4}{=}{\centering\arraybackslash \begin{minipage}[t]{\linewidth}\centering
Qwen2.5-Omni\\
(64.2) {[}7{]}\strut
\end{minipage}} & 65.95 & 66.07 & 65.82 & 64.99 \\
\multicolumn{2}{@{}>{\centering\arraybackslash}m{(\linewidth - 12\tabcolsep) * \real{0.2260} + 2\tabcolsep}}{%
GLM-4.5} & & 65.51 & 65.86 & 65.31 & 65.31 \\
\multicolumn{2}{@{}>{\centering\arraybackslash}m{(\linewidth - 12\tabcolsep) * \real{0.2260} + 2\tabcolsep}}{%
Doubao-Seed-1.6} & & 65.67 & 65.6 & 65.45 & 65.3 \\
\multicolumn{2}{@{}>{\centering\arraybackslash}m{(\linewidth - 12\tabcolsep) * \real{0.2260} + 2\tabcolsep}}{%
Qwen3} & & 65.67 & 65.83 & 65.45 & 65.46 \\
\multicolumn{2}{@{}>{\centering\arraybackslash}m{(\linewidth - 12\tabcolsep) * \real{0.2260} + 2\tabcolsep}}{%
GPT-4o} &
\multirow{4}{=}{\centering\arraybackslash \begin{minipage}[t]{\linewidth}\centering
Qwen2.5-VL\\
(61.68)\strut
\end{minipage}} & 62.72 & 62.75 & 62.64 & 62.14 \\
\multicolumn{2}{@{}>{\centering\arraybackslash}m{(\linewidth - 12\tabcolsep) * \real{0.2260} + 2\tabcolsep}}{%
GLM-4.5} & & 63.12 & 63.25 & 62.69 & 62.4 \\
\multicolumn{2}{@{}>{\centering\arraybackslash}m{(\linewidth - 12\tabcolsep) * \real{0.2260} + 2\tabcolsep}}{%
Doubao-Seed-1.6} & & 63.2 & 63.02 & 62.9 & 62.09 \\
\multicolumn{2}{@{}>{\centering\arraybackslash}m{(\linewidth - 12\tabcolsep) * \real{0.2260} + 2\tabcolsep}}{%
Qwen3} & & 63.81 & 63.83 & 63.78 & 63.39 \\
\end{longtable}

\textbf{Table 4.} WorldSense dataset: accuracy (\%) across combinations
of reasoning models and Pipeline module models

\begin{longtable}[]{@{}
  >{\centering\arraybackslash}m{(\linewidth - 12\tabcolsep) * \real{0.2144}}
  >{\centering\arraybackslash}m{(\linewidth - 12\tabcolsep) * \real{0.0116}}
  >{\centering\arraybackslash}m{(\linewidth - 12\tabcolsep) * \real{0.2023}}
  >{\centering\arraybackslash}m{(\linewidth - 12\tabcolsep) * \real{0.1429}}
  >{\centering\arraybackslash}m{(\linewidth - 12\tabcolsep) * \real{0.1429}}
  >{\centering\arraybackslash}m{(\linewidth - 12\tabcolsep) * \real{0.1429}}
  >{\centering\arraybackslash}m{(\linewidth - 12\tabcolsep) * \real{0.1429}}@{}}
\toprule\noalign{}
\begin{minipage}[b]{\linewidth}\centering
Pipeline
\end{minipage} &
\multicolumn{2}{>{\centering\arraybackslash}m{(\linewidth - 12\tabcolsep) * \real{0.2139} + 2\tabcolsep}}{%
\begin{minipage}[b]{\linewidth}\centering
Reasoning Model

(Baseline)
\end{minipage}} & \begin{minipage}[b]{\linewidth}\centering
CG\_Qwen

\end{minipage} & \begin{minipage}[b]{\linewidth}\centering
CG\_GLM

\end{minipage} & \begin{minipage}[b]{\linewidth}\centering
Abl\_NI

\end{minipage} & \begin{minipage}[b]{\linewidth}\centering
Abl\_SP

\end{minipage} \\
\midrule\noalign{}
\endhead
\bottomrule\noalign{}
\endlastfoot
\multicolumn{2}{@{}>{\centering\arraybackslash}m{(\linewidth - 12\tabcolsep) * \real{0.2260} + 2\tabcolsep}}{%
GPT-4o} &
\multirow{4}{=}{\centering\arraybackslash \begin{minipage}[t]{\linewidth}\centering
HumanOmniV2\\
(47.1) {[}7{]}\strut
\end{minipage}} & 48.8 & 48.55 & 47.79 & 48.14 \\
\multicolumn{2}{@{}>{\centering\arraybackslash}m{(\linewidth - 12\tabcolsep) * \real{0.2260} + 2\tabcolsep}}{%
GLM-4.5} & & 48.17 & 48.7 & 48.01 & 47.89 \\
\multicolumn{2}{@{}>{\centering\arraybackslash}m{(\linewidth - 12\tabcolsep) * \real{0.2260} + 2\tabcolsep}}{%
Doubao-Seed-1.6} & & 48.23 & 48.2 & 48.14 & 47.64 \\
\multicolumn{2}{@{}>{\centering\arraybackslash}m{(\linewidth - 12\tabcolsep) * \real{0.2260} + 2\tabcolsep}}{%
Qwen3} & & 48.36 & 48.36 & 48.3 & 47.92 \\
\multicolumn{2}{@{}>{\centering\arraybackslash}m{(\linewidth - 12\tabcolsep) * \real{0.2260} + 2\tabcolsep}}{%
GPT-4o} &
\multirow{4}{=}{\centering\arraybackslash \begin{minipage}[t]{\linewidth}\centering
Qwen2.5-Omni\\
(45.4) {[}7{]}\strut
\end{minipage}} & 47.13 & 47.67 & 47.01 & 46.75 \\
\multicolumn{2}{@{}>{\centering\arraybackslash}m{(\linewidth - 12\tabcolsep) * \real{0.2260} + 2\tabcolsep}}{%
GLM-4.5} & & 47.57 & 47.38 & 47.23 & 46.31 \\
\multicolumn{2}{@{}>{\centering\arraybackslash}m{(\linewidth - 12\tabcolsep) * \real{0.2260} + 2\tabcolsep}}{%
Doubao-Seed-1.6} & & 47.45 & 47.04 & 46.94 & 46.69 \\
\multicolumn{2}{@{}>{\centering\arraybackslash}m{(\linewidth - 12\tabcolsep) * \real{0.2260} + 2\tabcolsep}}{%
Qwen3} & & 47.86 & 47.79 & 47.7 & 46.22 \\
\multicolumn{2}{@{}>{\centering\arraybackslash}m{(\linewidth - 12\tabcolsep) * \real{0.2260} + 2\tabcolsep}}{%
GPT-4o} &
\multirow{4}{=}{\centering\arraybackslash \begin{minipage}[t]{\linewidth}\centering
Qwen2.5-VL\\
(37.39)\strut
\end{minipage}} & 43.1 & 43.1 & 42.97 & 41.87 \\
\multicolumn{2}{@{}>{\centering\arraybackslash}m{(\linewidth - 12\tabcolsep) * \real{0.2260} + 2\tabcolsep}}{%
GLM-4.5} & & 43.41 & 42.88 & 42.4 & 41.93 \\
\multicolumn{2}{@{}>{\centering\arraybackslash}m{(\linewidth - 12\tabcolsep) * \real{0.2260} + 2\tabcolsep}}{%
Doubao-Seed-1.6} & & 42.21 & 42.12 & 41.93 & 41.33 \\
\multicolumn{2}{@{}>{\centering\arraybackslash}m{(\linewidth - 12\tabcolsep) * \real{0.2260} + 2\tabcolsep}}{%
Qwen3} & & 43.06 & 43.25 & 42.91 & 41.83 \\
\end{longtable}

\textbf{Table 5.} Daily-Omni dataset: accuracy (\%) across combinations
of reasoning models and Pipeline module models

\begin{longtable}[]{@{}
  >{\centering\arraybackslash}m{(\linewidth - 12\tabcolsep) * \real{0.2144}}
  >{\centering\arraybackslash}m{(\linewidth - 12\tabcolsep) * \real{0.0116}}
  >{\centering\arraybackslash}m{(\linewidth - 12\tabcolsep) * \real{0.2023}}
  >{\centering\arraybackslash}m{(\linewidth - 12\tabcolsep) * \real{0.1429}}
  >{\centering\arraybackslash}m{(\linewidth - 12\tabcolsep) * \real{0.1429}}
  >{\centering\arraybackslash}m{(\linewidth - 12\tabcolsep) * \real{0.1429}}
  >{\centering\arraybackslash}m{(\linewidth - 12\tabcolsep) * \real{0.1429}}@{}}
\toprule\noalign{}
\begin{minipage}[b]{\linewidth}\centering
Pipeline
\end{minipage} &
\multicolumn{2}{>{\centering\arraybackslash}m{(\linewidth - 12\tabcolsep) * \real{0.2139} + 2\tabcolsep}}{%
\begin{minipage}[b]{\linewidth}\centering
Reasoning Model

(Baseline)
\end{minipage}} & \begin{minipage}[b]{\linewidth}\centering
CG\_Qwen

\end{minipage} & \begin{minipage}[b]{\linewidth}\centering
CG\_GLM

\end{minipage} & \begin{minipage}[b]{\linewidth}\centering
Abl\_NI

\end{minipage} & \begin{minipage}[b]{\linewidth}\centering
Abl\_SP

\end{minipage} \\
\midrule\noalign{}
\endhead
\bottomrule\noalign{}
\endlastfoot
\multicolumn{2}{@{}>{\centering\arraybackslash}m{(\linewidth - 12\tabcolsep) * \real{0.2260} + 2\tabcolsep}}{%
GPT-4o} &
\multirow{4}{=}{\centering\arraybackslash \begin{minipage}[t]{\linewidth}\centering
HumanOmniV2\\
(58.47) {[}7{]}\strut
\end{minipage}} & 60.23 & 59.9 & 58.56 & 59.31 \\
\multicolumn{2}{@{}>{\centering\arraybackslash}m{(\linewidth - 12\tabcolsep) * \real{0.2260} + 2\tabcolsep}}{%
GLM-4.5} & & 62.74 & 61.24 & 59.31 & 60.74 \\
\multicolumn{2}{@{}>{\centering\arraybackslash}m{(\linewidth - 12\tabcolsep) * \real{0.2260} + 2\tabcolsep}}{%
Doubao-Seed-1.6} & & 62.66 & 61.07 & 59.9 & 60.9 \\
\multicolumn{2}{@{}>{\centering\arraybackslash}m{(\linewidth - 12\tabcolsep) * \real{0.2260} + 2\tabcolsep}}{%
Qwen3} & & 62.49 & 61.32 & 61.24 & 60.15 \\
\multicolumn{2}{@{}>{\centering\arraybackslash}m{(\linewidth - 12\tabcolsep) * \real{0.2260} + 2\tabcolsep}}{%
GPT-4o} &
\multirow{4}{=}{\centering\arraybackslash \begin{minipage}[t]{\linewidth}\centering
Qwen2.5-Omni\\
(47.45) {[}7{]}\strut
\end{minipage}} & 55.56 & 55.64 & 55.47 & 50.38 \\
\multicolumn{2}{@{}>{\centering\arraybackslash}m{(\linewidth - 12\tabcolsep) * \real{0.2260} + 2\tabcolsep}}{%
GLM-4.5} & & 54.05 & 54.22 & 53.38 & 50.88 \\
\multicolumn{2}{@{}>{\centering\arraybackslash}m{(\linewidth - 12\tabcolsep) * \real{0.2260} + 2\tabcolsep}}{%
Doubao-Seed-1.6} & & 54.89 & 52.38 & 51.88 & 50.54 \\
\multicolumn{2}{@{}>{\centering\arraybackslash}m{(\linewidth - 12\tabcolsep) * \real{0.2260} + 2\tabcolsep}}{%
Qwen3} & & 56.9 & 56.96 & 56.81 & 51.88 \\
\multicolumn{2}{@{}>{\centering\arraybackslash}m{(\linewidth - 12\tabcolsep) * \real{0.2260} + 2\tabcolsep}}{%
GPT-4o} &
\multirow{4}{=}{\centering\arraybackslash \begin{minipage}[t]{\linewidth}\centering
Qwen2.5-VL\\
(47.28)\strut
\end{minipage}} & 49.71 & 49.96 & 49.62 & 49.12 \\
\multicolumn{2}{@{}>{\centering\arraybackslash}m{(\linewidth - 12\tabcolsep) * \real{0.2260} + 2\tabcolsep}}{%
GLM-4.5} & & 51.55 & 51.71 & 51.38 & 50.96 \\
\multicolumn{2}{@{}>{\centering\arraybackslash}m{(\linewidth - 12\tabcolsep) * \real{0.2260} + 2\tabcolsep}}{%
Doubao-Seed-1.6} & & 50.79 & 50.63 & 50.46 & 50.35 \\
\multicolumn{2}{@{}>{\centering\arraybackslash}m{(\linewidth - 12\tabcolsep) * \real{0.2260} + 2\tabcolsep}}{%
Qwen3} & & 51.63 & 51.55 & 51.46 & 50.54 \\
\end{longtable}

From Table 3, on the IntentBench dataset: for HumanOmniV2, the baseline accuracy is 69.33\%, and with our method it reaches a maximum of 71.18\% (+1.85 pp; CG\_Qwen; Pipeline = Qwen3). For Qwen2.5-Omni, the baseline is 64.20\%, with a maximum of 66.07\% (+1.87 pp; CG\_GLM; Pipeline = GPT-4o). For Qwen2.5-VL, the baseline is 61.68\%, with a maximum of 63.83\% (+2.15 pp; CG\_GLM; Pipeline = Qwen3).

On the WorldSense dataset (Table 4): HumanOmniV2 baseline 47.10\%, maximum 48.80\% (+1.70 pp; CG\_Qwen; Pipeline = GPT-4o); Qwen2.5-Omni baseline 45.40\%, maximum 47.86\% (+2.46 pp; CG\_Qwen; Pipeline = Qwen3); Qwen2.5-VL baseline 37.39\%, maximum 43.41\% (+6.02 pp; CG\_Qwen; Pipeline = GLM-4.5).

Similarly, on the Daily-Omni dataset (Table 5): HumanOmniV2 baseline 58.47\%, maximum 62.74\% (+4.27 pp; CG\_Qwen; Pipeline = GLM-4.5); Qwen2.5-Omni baseline 47.45\%, maximum 56.96\% (+9.51 pp; CG\_GLM; Pipeline = Qwen3); Qwen2.5-VL baseline 47.28\%, maximum 51.71\% (+4.43 pp; CG\_GLM; Pipeline = GLM-4.5).

Table 3-5 report detailed results on three benchmarks, respectively. The best results are sometimes achieved by CG\_Qwen and sometimes by the CG\_GLM, but both surpass ``no intent'' (Abl\_NI) and ``single policy'' (Abl\_SP). In addition, both Abl\_NI and Abl\_SP outperform the baseline, confirming the effectiveness of each module. The maximum gain corresponds to +9.51 pp (a 20.04\% relative improvement). Importantly, the three-module scheme consistently outperforms the corresponding baselines across all combinations of the four pipelines and three reasoning engines, regardless of whether the pipeline LLMs are closed-source or open-source, this indicates that our method offers strong portability and plug-and-play characteristics.

\subsection{3.3 Ablation Study and
Analyses}\label{ablation-study-and-analyses}

We ablate the three modules---Intent Perceiver (IP), Policy Generator (PG), and Strategy Selector (SS)---with results summarized in Table~III. Removing any single component degrades accuracy. For clarity, we denote Abl\_NI as removing IP (i.e., no \(Z_{\mathrm{IP}}\)), and Abl\_SP as using a single policy (\(N{=}1\)), which disables PG’s diversity and makes SS degenerate to selecting the only sketch. For Daily-Omni (HumanOmniV2 \(\times\) GLM-4.5), the full system reaches \(62.74\%\); Abl\_NI drops to \(59.31\%\) (\(-3.43\) pp), and Abl\_SP to \(60.74\%\) (\(-2.00\) pp). Similar patterns hold elsewhere: on WorldSense (Qwen2.5-VL \(\times\) GLM-4.5) full \(43.41\%\), no-IP \(42.40\%\) (\(-1.01\) pp), single-policy \(41.93\%\) (\(-1.48\) pp); on IntentBench (HumanOmniV2 \(\times\) Qwen3) full \(71.18\%\), no-IP \(70.82\%\) (\(-0.36\) pp), single-policy \(69.96\%\) (\(-1.22\) pp).

Interpretation. Without the intent provided by the Intent Perceiver, the model may miss crucial omni-modal cues, typically resulting in a modest performance drop; with Abl\_SP (\(N{=}1\)), the strategy set collapses and coverage over semantic classes \(\mathcal{M}\) shrinks, which aligns with the increase of semantic uncertainty in Eqs.~(2)--(3). Note that Abl\_SP does not remove SS; rather, the selector trivially chooses the only available sketch. Therefore, the observed decline under Abl\_SP should be attributed to the loss of PG-induced diversity (and the \(\mathrm{Div}(\cdot)\) term in Eq.~(5)). Among the ablations, the impact of policy generation is larger: the single-policy (Abl\_SP) degradation tends to exceed the no-IP case, indicating that multi-path thinking is intrinsically valuable. This mirrors the human ``understand--plan--select'' pattern, where careful multi-path simulation precedes a final decision. The selector’s role is captured by minimum conditional entropy/Bayesian risk (Eqs.~(6)--(9)); when \(N{>}1\) and SS actively scores candidate sketches, uncertainty is further reduced in line with the contraction in Eqs.~(11)--(13). From the \emph{pipeline} perspective, Qwen3-based pipelines exhibit the highest win rate across settings; given that most reasoning engines here are from the Qwen family, this ``same-lineage'' pairing likely improves adherence to strategy guidance and reduces mismatch between policy sketches and the executor. From the \emph{reasoning-model} perspective, omni models outperform VL models, suggesting that access to audio cues materially improves disambiguation and intent grounding.

Our approach is model- and framework-agnostic. Across reasoning engines (HumanOmniV2, Qwen2.5-Omni, Qwen2.5-VL) and pipelines (GPT-4o / GLM-4.5 / Doubao-Seed-1.6 / Qwen3), improvements are consistent. Gains are larger for base models lacking post-training (e.g., WorldSense: Qwen2.5-VL \(37.39\%\rightarrow 43.41\%\), \(+6.02\) pp, CG\_Qwen, Pipeline{=}GLM-4.5; HumanOmniV2 \(47.10\%\rightarrow 48.80\%\), \(+1.70\) pp, Pipeline{=}GPT-4o). On Daily-Omni, HumanOmniV2 improves \(58.47\%\rightarrow 62.74\%\) (\(+4.27\) pp) with GLM-4.5; replacing GLM-4.5 with the smaller Qwen3 still yields \(+4.02\) pp. These results indicate a plug-and-play component that reliably enhances reasoning across open/closed models and scales, demonstrating cross-model, cross-platform, and cross-scenario robustness.

\section{\texorpdfstring{ CONCLUSION}{ CONCLUSION}}\label{conclusion}

This paper presents a zero-shot omni-modal reasoning component that implements a plug-and-play, intent-sketch pipeline: Intent Perceiver, Policy Generator, and Strategy Selector, which uses context injection to enhance reasoning performance without fine-tuning and generalizes across modalities, reasoning engines, and platforms. From an information-theoretic view, Policy Generator produces complementary candidates conditioned on intent, Strategy Selector selects among candidate strategies via prompt-based evaluation under additional task-specific conditioning, effectively contracting the answer posterior and reducing decision uncertainty, explaining the stable gains and improved interpretability without training overhead.

On IntentBench, WorldSense, and Daily-Omni, the full three-module scheme consistently outperforms baselines across all Pipeline/engine combinations, with maximum gains of +9.51 pp (relative improvement of 20.04\%). Ablations show complementary roles: PG+SS is the primary source of improvement, while explicit intent is especially beneficial for video/audio-dependent tasks. The method delivers stable cross-platform gains for diverse reasoning engines and Pipeline models, underscoring its plug-and-play nature and strong transferability. Overall, the intent sketch offers a lightweight, effective paradigm for improving the alignment, robustness, and interpretability of complex cross-modal reasoning.

\section{\texorpdfstring{ REFERENCES}{ REFERENCES}}\label{references}

{[}1{]} S. Bubeck, V. Chandrasekaran, R. Eldan, J. Gehrke, E. Horvitz,
E. Kamar, P. Lee, Y. T. Lee, Y. Li, S. Lundberg, H. Nori, H. Palangi, M.
T. Ribeiro, and Y. Zhang, ``Sparks of Artificial General Intelligence:
Early experiments with GPT-4,'' \emph{arXiv preprint arXiv:2303.12712},
2023.

{[}2{]} D. Driess, F. Xia, M. S. M. Sajjadi, C. Lynch, A. Chowdhery, B.
Ichter, A. Wahid, J. Tompson, Q. Vuong, T. Yu, W. Huang, Y. Chebotar, P.
Sermanet, D. Duckworth, S. Levine, V. Vanhoucke, K. Hausman, M.
Toussaint, K. Greff, A. Zeng, I. Mordatch, and P. Florence, ``PaLM-E: An
Embodied Multimodal Language Model,'' in \emph{ICML}, 2023, pp.
8469--8488.

{[}3{]} Z. Yan, Z. Li, Y. He, C. Wang, K. Li, X. Li, X. Zeng, Z. Wang,
Y. Wang, Y. Qiao, L. Wang, and Y. Wang, ``Task Preference Optimization:
Improving Multimodal Large Language Models with Vision Task Alignment,''
in \emph{CVPR}, 2025, pp. 29880--29892.

{[}4{]} J. Li, P. Wei, W. Han, and L. Fan, ``IntentQA: Context-aware
Video Intent Reasoning,'' in \emph{ICCV}, 2023, pp. 11963--11974

{[}5{]} M. Ma, Z. Yu, Y. Ma, G. Li and Z. Yang, "EventLens: Enhancing
Visual Commonsense Reasoning by Leveraging Event-Aware Pretraining and
Cross-modal Linking,"~in \emph{ICASSP 2025 - 2025 IEEE International
Conference on Acoustics, Speech and Signal Processing (ICASSP)}, IEEE,
2025, pp. 1-5

{[}6{]} D. Zhang, Y. Yu, J. Dong, C. Li, D. Su, C. Chu, and D. Yu,
``MM-LLMs: Recent Advances in MultiModal Large Language Models,'' in
ACL, 2024, pp. 12401--12430.

{[}7{]} Q. Yang, S. Yao, W. Chen, S. Fu, D. Bai, J. Zhao, B. Sun, B.
Yin, X. Wei, and J. Zhou, ``HumanOmniV2: From Understanding to
Omni-Modal Reasoning with Context,'' \emph{arXiv preprint
arXiv:2506.21277}, 2025.

{[}8{]} X. Zheng, C. Liao, Y. Fu, K. Lei, Y. Lyu, L. Jiang, B. Ren, J.
Chen, J. Wang, C. Li, L. Zhang, D. P. Paudel, X. Huang, Y.-G. Jiang, N.
Sebe, D. Tao, L. V. Gool, and X. Hu, ``MLLMs are Deeply Affected by
Modality Bias,'' \emph{arXiv preprint arXiv:2505.18657}, 2025.

{[}9{]} Y. Hao, J. Gu, H. W. Wang, L. Li, Z. Yang, L. Wang, and Y.
Cheng, ``Can MLLMs Reason in Multimodality? EMMA: An Enhanced MultiModal
ReAsoning Benchmark,'' in \emph{ICML}, 2025.

{[}10{]} J. Xia, Y. Zang, P. Gao, Y. Li, and K. Zhou, ``Visionary-R1:
Mitigating Shortcuts in Visual Reasoning with Reinforcement Learning,''
arXiv preprint \emph{arXiv:2505.14677}, 2025.

{[}11{]} Z. Hu, A. Iscen, C. Sun, Z. Wang, K.-W. Chang, Y. Sun, C.
Schmid, D. A. Ross, and A. Fathi, ``REVEAL: Retrieval-Augmented
Visual-Language Pre-Training with Multi-Source Multimodal Knowledge
Memory,'' in \emph{CVPR}, 2023, pp. 23369--23379.

{[}12{]} S. Ravi, A. Chinchure, L. Sigal, R. Liao, and V. Shwartz,
``VLC-BERT: Visual Question Answering with Contextualized Commonsense
Knowledge,'' in \emph{WACV}, 2023, pp. 1155--1165.

{[}13{]} Y. Zhang, B. Chen, H. Ye, Z. Gao, T. Wan and L. Lan,
"Text-guided Multimodal Fusion for the Multimodal Emotion and Intent
Joint Understanding," in \emph{ICASSP 2025 - 2025 IEEE International
Conference on Acoustics, Speech and Signal Processing (ICASSP)}, IEEE,
2025, pp. 1-2

{[}14{]} J. Hong, S. Yan, J. Cai, X. Jiang, Y. Hu, and W. Xie,
``WorldSense: Evaluating Real-world Omnimodal Understanding for
Multimodal LLMs,'' \emph{arXiv preprint arXiv:2502.04326}, 2025.

{[}15{]} Z. Zhou, R. Wang, and Z. Wu, ``Daily-Omni: Towards Audio-Visual
Reasoning with Temporal Alignment across Modalities,'' \emph{arXiv
preprint arXiv:2505.17862}, 2025.

{[}16{]} Z. Zhang, A. Zhang, M. Li, H. Zhao, G. Karypis, and A. Smola,
``Multimodal Chain-of-Thought Reasoning in Language Models,'' in
\emph{TMLR}, 2024.

{[}17{]} S. A. Aytes, J. Baek, and S. J. Hwang, ``Sketch-of-Thought:
Efficient LLM Reasoning with Adaptive Cognitive-Inspired Sketching,''
\emph{arXiv preprint arXiv:2503.05179}, 2025.

{[}18{]} Z. Yang, X. Yu, D. Chen, M. Shen, and C. Gan, ``Machine Mental
Imagery: Empower Multimodal Reasoning with Latent Visual Tokens,''
\emph{arXiv preprint arXiv:2506.17218}, 2025.

\end{document}